# Efficient Object Detection of Marine Debris using Pruned YOLO Model

Abi Aryaza[a], Novanto Yudistira[a], Tibyani[a]

[a]*Informatics Department, Faculty of Computer Science, Brawijaya University, 65145, Malang, Indonesia*

**Abstract**

Marine debris remains a persistent and crucial issue that demands attention, as the substances present in the waste pose significant harm to marine life. Ingredients such as microplastics, polychlorinated biphenyls, and pesticides can poison and damage the habitats of organisms living in proximity. However, solutions involving human labor, such as diving, are becoming increasingly ineffective due to the limitations of humans in underwater environments. To address this challenge, technology involving autonomous underwater vehicles continues to be developed for effective sea garbage collection. In this development process, the selection of object detection architecture in autonomous underwater vehicles plays a critical role. For the creation of robots capable of handling marine debris, a one-stage detector type architecture is highly recommended due to its necessity for real-time detection. This research focuses on utilizing the You Only Look Once model version 4 (YOLOv4) architecture for the object detection of marine debris. The dataset utilized in this research comprises 7683 images of marine debris collected in Trash-ICRA 19 dataset with 480x320 pixels. Furthermore, various modifications, such as using pretrained model, training from scratch, disable mosaic augmentation, enable mosaic augmentation, freezing the backbone only layer, freezing the backbone and neck layer, YOLOv4-tiny, and adding channel pruning of YOLOv4, are implemented each other and compared to find the most impactful for enhance the efficiency of the architecture. Numerous studies have demonstrated that the application of channel pruning can improve detection speed without significantly sacrificing accuracy. The dataset employed in this research is the trash-ICRA 19 dataset, featuring images of objects in seawater. This dataset emphasizes images of plastic waste objects in water, incorporating several other classes. Through the application of channel pruning to YOLOv4 trained on the underwater object image dataset, the frame rate per second value from the base YOLOv4 is from 15.19FPS increases to 19.4 FPS, accompanied by only 1.2% reduction in mean Average Precision from 97.6% to 96.4%.

*Keywords: Marine Debris, Object Detection, YOLO, Autonomous Underwater Vehicles, Channel Pruning*

## 1. Introduction

Environmental pollution has always been a serious problem worldwide, and one contributing factor stems from waste—the remnants or traces of everything discarded by humans[1]. In 2022, the generation of waste in Indonesia reached 35 million tons, with 23.5 million tons being managed, and 11.6 million tons left unmanaged[2]. Unmanaged waste can occur for several reasons, such as few landfills, and limited waste processing capacity. As a result, unmanaged waste is dispersed in various places, such as buried in the ground or flowing into the sea. All waste disposed of or abandoned in marine environments or the Great Lakes, whether intentionally or not, is termed marine debris[3]. Marine debris encompasses nonbiodegradable solids, manufactured or processed products, originating from diverse sources like fishing activities, beach tourism, industrial waste, and improper garbage disposal on land. Classifications of marine debris include plastic, metal, glass, rubber, and organic materials. Approximately 14 billion tons of waste find their way into the ocean each year[4], with plastic waste dominating as the most prevalent type, constituting around 60-80 percent of total marine debris[5].

Currently, robots and artificial intelligence play crucial roles in various research endeavors. In underwater research, robots are essential for observing environments that are challenging for humans to reach, such as areas with limited water depth, leak detection in underwater pipelines, and more. Autonomous Underwater Vehicles (AUVs) equipped with sensors like side-scan sonar, cameras, echosounders, Acoustic Doppler Current Profilers (ADCP), and Conductivity, Temperature, and Depth (CTD) sensors prove highly useful for water observation[6]. AUVs, being unmanned, are automatically controlled by a computer[7].

One of the focal points in artificial intelligence is computer vision, which involves developing algorithms capable of understanding and extracting information from images and videos, mirroring the way humans use vision to perceive information[8]. Computer vision encompasses technologies like digital image processing, pattern recognition, and data analysis techniques such as machine learning and deep learning[9]. Object detection, a case study in image processing and computer vision, aims to identify the presence and position of specific objects in images or videos. Object detection has evolved rapidly since the introduction of the Region-based Convolutional Neural Networks (R-CNN) mode in 2014[10]. Numerous models have since been created to address tasks related to object recognition and localization, categorizing object detection into two types: two-stage detectors and one-stage detectors. Two-stage detectors perform detections through two stages: region proposal and classification. While this makes detection more reliable, it increases computational costs due to complexity[11]. Conversely, one-stage detectors use a single feed-forward network, making them faster and suitable for real-time applications, albeit with slightly less reliable detection results[12]. Object detection finds applications in various fields, including defense[13], facial recognition[14], transport[15], robots[16], personal protective equipment safety[17], medicine[18], and environmental monitoring.

Research on object detection underwater presents unique challenges. Detection underwater remains challenging due to the environment where water causes 'light attenuation,' leading to the absorption and scattering of light[19]. Another challenge is the diversity of objects in the ocean, making it harder to detect specific objects. Images and videos of objects in the ocean are not as easily obtained as those on land, given human limitations in capturing oceanic visuals. However, focusing on detecting plastic, which constitutes around 60-80 percent of total marine debris[5], can simplify the task. Plastic also poses a significant threat to marine ecosystems.

The dataset utilized in this research is Trash-ICRA 19, a comprehensive collection enhancing object detection capabilities[28]. Comprising 7683 images, each with dimensions of 480x320 pixels, the dataset is meticulously organized for effective training and evaluation. It is strategically divided into 5719 training images, 1144 test images, and 820 validation images, facilitating robust model development and assessment.

Annotating the dataset precisely, three distinct classes have been identified to enrich the object detection process: "Plastic" for plastic material associated with marine debris; "ROV" for components of Remotely Operated Vehicles; and "Living things in the water" for various living organisms in aquatic environments. This detailed annotation refines the training process and enables the model to discern and categorize objects with heightened accuracy. The Trash-ICRA 19 dataset serves as a pivotal asset in advancing research on marine debris detection and cleanup. Capturing images every 3 frames per second results in many similar images, and to overcome this, dataset transformations such as random cropping, color jitter, scale rotation, affine transformation, blur, and others are performed. Figure 1 shows several images after applying these transformations.

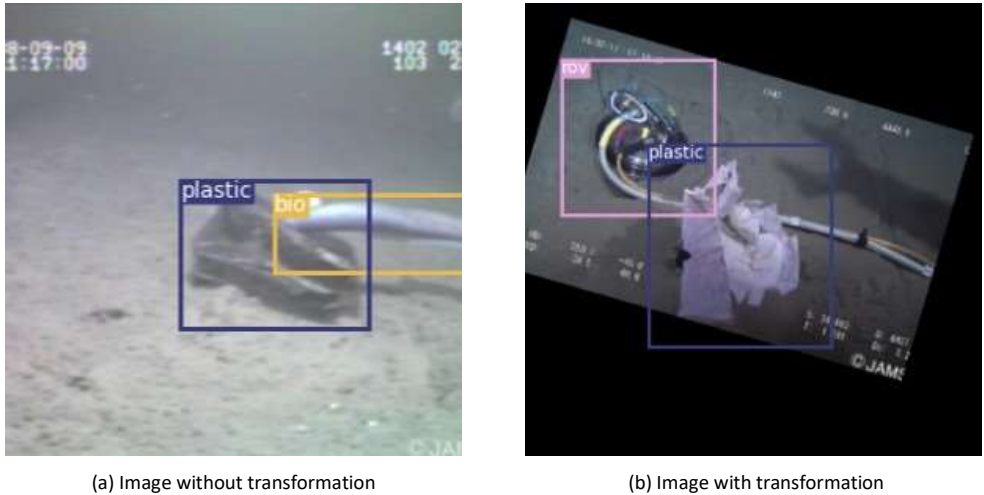

(a) Image without transformation      (b) Image with transformation

Figure 1: Example of image dataset

Given the two types of object detection, opting for a one-stage detector is more suitable for Autonomous Underwater Vehicles (AUVs). AUVs are designed for the purpose of collecting trash in the ocean, necessitating real-time detection capabilities. YOLO[20] [21] and Single-Shot Detector (SSD)[22] [23] are several of notable object detection models that can be employed for real-time object detection underwater. In this study, YOLO, a one-stage detector that is notable for its speed and capacity for rapid detection, is employed on the Trash-ICRA 19 dataset. YOLOv4, representing the current state of the art in object detection, is selected in preference to other one-stage detectors such as SSD due to its enhanced speed and accuracy in detection. The implementation of the backbone layer utilizing CSPDarknet53 and the neck employing PANet facilitates enhanced feature extraction, thereby optimizing the accuracy and speed of detection. [24][25]. YOLOv4 has exhibited remarkable efficacy in a diverse range of object detection tasks, including those involving underwater objects[26][27][21]. Several studies have been conducted related to object detection using YOLO. Focused on YOLOv4, previous research indicates that the mosaic augmentation method, one of the YOLOv4 innovations, produces lower values than without the method[17]. Additionally, it is interesting to investigate the impact of fine-tuning a pretrained model with a frozen layer. Moreover, a lightweight version of YOLOv4, designated as YOLOv4-tiny, has been developed. instead using YOLOv4-tiny, a 'pruning' method also used to increase speed without sacrificing too much accuracy. Because of this curiosity the authors successfully make several contributions to the field.

In summary, the three main contributions of this study are as follows:

・ Developed a one-stage detector model, YOLOv4, that focuses on improving efficiency and scalability specific to the marine debris detection problem.

- Successfully implements the channel-pruning technique in YOLOv4 with marine debris dataset that make the computational cost reduced without decreasing accuracy, making the model better and make it available for being used even by low embedded computing system.

- The proposed efficient YOLO model demonstrates an increase in frames per second (FPS) of the baseline model from 15.19 FPS to 19.4 FPS when detect plastic waste that indicate the proposed model is applicable for real-world marine debris.

## 2. Related Works

Fulton et al., 2019 [28], conducted a study entitled 'Robotic Detection of Marine Litter Using Deep Visual Detection Models,' which served as the reference for the dataset used in this study. The YOLOv2 method was employed with a mean average precision (mAP) value of 47.9%, achieving an frames per second (FPS) of 205 using GTX 1080Ti.

Yang et al., 2021 [20], conducted a study entitled 'Research on Underwater Object Recognition based on YOLOv3' on the same dataset, utilizing the YOLOv3 method. They achieved a mAP of 76.1%, recall of 75.6%, with a frame rate of 20 FPS. The study also included a comparison between the YOLOv3 method and Faster R-CNN[46].

Tian et al., 2022 [27], conducted a study entitled 'A modified YOLOv4 detection method for a vision-based underwater garbage cleaning robot' on the same topic, focusing on marine garbage using the YOLOv4 method. The research aimed to detect objects by comparing several methods, such as 4SP- YOLOv4, 4S-YOLOv4, YOLOv4, YOLOv3, Faster R-CNN, and SSD. The 4S-YOLOv4 method obtained the best mAP value at 95.5%. This method, a modified YOLOv4 version, successfully increased the accuracy by introducing a 4-scale YOLOv4. Meanwhile, the best FPS was achieved by the 4SP-YOLOv4 method by trimming the 4-scale YOLOv4 version, reducing parameters, and optimizing models.

Majchrowska et al., 2021 [29], conducted a study entitled 'Waste detection in Pomerania: a non-profit project for detecting waste in the environment,' focusing on waste detection. The research involved various datasets, training, and dataset combination. The Trash-ICRA 19 dataset was one of the tested datasets, with evaluation results in the study showing a mAP value of only 7.3% with the EfficientDet-D2 model using the EfficientNet-B2 backbone. According to the researchers, poor image quality resulted in mAP values below 10%. Furthermore, the dataset was combined with other datasets and trained, achieving a final accuracy of 73.02% using Weighted Sampler to produce a more stable evaluation of each class.

Liao Juang, 2023 [21], conducted a study entitled 'Automatic Marine Debris

Inspection,' with a topic similar to this study, specifically litter detection. The study utilized the HAIDA dataset, consisting of 2 classes, namely 'garbage' and 'bottles.' The model used was YOLOv4-tiny with CSPOSANet backbone, achieving AP50 results on testing of 73.37%.

S´anchez-Ferrer et al., 2023 [30], conducted a study entitled 'An experimental study on marine debris location and recognition using object detection,' focusing on underwater debris detection and segmentation. The dataset used was a collection of datasets from CleanSea and JAMSTEC, supplemented by synthetic data to overcome data limitations. For detection and segmentation, the study employed the Mask R-CNN model with mixed results. With a data division of 50% for real data and synthetic data, the resulting mAP50 value was 52.1%, while without synthetic data, the mAP50 value reached 61

Tian et al., 2021 [31], conducted a study entitled 'Pruning-Based YOLOv4 Algorithm for Underwater Garbage Detection,' focusing on garbage detection in water. The model used in this research was YOLOv4, incorporating channel pruning and layer pruning. The results obtained for the YOLOv4 model were a mAP of 91.3% with 43.4 FPS, while after pruning, the mAP was 90.3% with 58.82 FPS.

Tata et al., 2021 [22], conducted a study entitled 'A Robotic Approach towards Quantifying Epipelagic Bound Plastic,' focusing on plastic waste detection in the sea. The research combined primary datasets and datasets from JAMSTEC, then published open-source with the dataset name DeepTrash. Four models were used in the study: SSD, Faster R-CNN, YOLOv4-tiny, and YOLOv5s. The evaluation results showed that the best mAP was obtained from the YOLOv5s model with an mAP value of 85%, a 1% difference from the YOLOv4-tiny model which achieved an mAP value of 84%. Inference time on YOLOv4-tiny reached 1.2 ms/img, while YOLOv5s reached
1.4 ms/img.

Xue et al., 2021[23] conducted a study entitled 'An Efficient Deep-Sea Debris Detection Method Using Deep Neural Networks' on the topic of marine debris detection. The research took images from JAMSTEC which were then manually labeled by the researcher, and the construction results were named the 3-D dataset. The number of datasets is around 10000 images with 7 labels. The research used several models namely R-CNN, SSD, and YOLOv3 with each model using the backbone namely MobileNetV2, Visual Geometry Group(VGG)16[43], and Residual Network(ResNet)50[49]. Based on the backbone, all models get the best Average Precision(AP)50 when using ResNet50. While overall of the 9 architectures built, the best AP50 value is obtained on the ResNet50-YOLOv3 architecture with AP values in each class being 61.7% (clothes), 86% (fishing line and rope), 91.6% (glass), 85.2% (iron), 82.5% (natural debris), 79.4% (plastic), and 97.6% (rubber).

Tajbakhsh et al., 2016[32] conducted a study entitled 'Convolutional Neural

Networks for Medical Image Analysis: Full Training or Fine Tuning?' on the topic of medical detection. The study used the Alex-Net [42] model with focus on the number of layers to freeze and the number of datasets to train. Using ROC curves, the results showed that fine-tuning the last half of the Alex-Net produced higher scores than fine-tuning the entire layer at 10% and 15% false positive rates. While fine-tuning only the last layer produced the lowest scores

Nugraha et al., 2016[17] conducted a study titled 'Supervised Virtual-To-Real Domain Adaptation for Object Detection Task Using YOLO.' The study aimed to perform object detection on Personal Protective Equipment (PPE) data using YOLOv4. The research implemented Source Hypothesis Transfer (SHOT) for the domain adaptation scheme. Additionally, a comparison testing scheme was conducted with and without mosaic augmentation. In the study, testing without mosaic augmentation resulted in a higher mAP of 65.5%, compared to testing with mosaic augmentation which obtained a lower mAP of only 55%.

## 3. Methodology

### 3.1. YOLO

The YOLO method presents a single-stage approach to object detection, employing an innovative technique. Object detection categorically falls into two types: single-stage detectors and two-stage detectors. In the case of single-stage detectors, the detection process concentrates on a sole evaluation network without traversing through the proposal region. YOLO stands out by utilizing a single evaluation network, resulting in faster detection speeds compared to other object detection methods[19]. However, this approach does affect the confidence level of the detection results, which tends to be lower than that of the two-stage detector.

The process initiates by converting the input image into a 3D matrix. Subsequent to the training process, each cell within a grid predicts N possible bounding boxes and confidence values[33]. Furthermore, bounding boxes with confidence values falling below the threshold are eliminated, retaining only those with high confidence levels. The bounding boxes are then arranged based on their confidence value and subjected to the non-maximum suppression (NMS) process. This process scrutinizes similar and adjacent bounding box layouts to acquire a singular bounding box representing an object[34]. The NMS calculation employs the intersection over union (IoU) metric. Specifically, the IoU between bounding box A and bounding box B is calculated, and if the result exceeds the confidence threshold, bounding box B is removed.

The development of detection objects continues to evolve over time until now, and for the evolution of detection objects until 2020 can be seen in Figure 2.

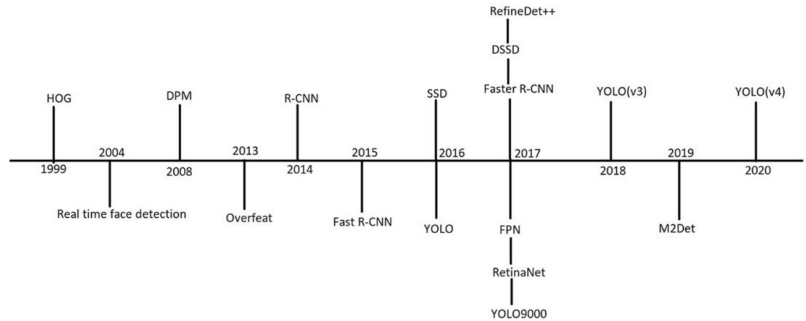

Figure 2: Object Detection Evolution[33]

YOLOv4 represents an advancement of YOLO, offering the flexibility to choose its backbone. By selecting CSPDarknet53 as the backbone, YOLOv4 achieves superior mAP and FPS compared to CSPResNeXt-50 and EfficientNet- B0[35]. CSPDarknet53 is an enhanced iteration of Darknet-53, the original backbone for YOLOv3. This enhancement involves the incorporation of CSPNet (Cross Stage Partial Network) into the residual block and a transition in the activation layer to include mish activation and Leaky-ReLU activation[36]. Mish, as an activation function, possesses self-adjusting and non-monotonic properties, resulting in smoother output compared to the ReLU activation function used in darknet-53[37].

Within the neck layer, YOLOv4 incorporates SPP (Spatial Pyramid Pooling) and PANet (Path Aggregation Network). SPP integrates a max-pooling layer with three pooling sizes: $5\times5$, $9\times9$, and $13\times1$[38]. While originally designed for instance segmentation, PANet, as employed in YOLOv4, conducts upsampling and downsampling using both low-level and high-level feature maps[39].

The YOLOv4 head mirrors the YOLOv3 head, featuring three outputs. The first output is tailored for detecting small objects, the second output for medium-sized objects, and the third output for large objects[40].

The constituent elements of YOLOv4 operate distinctly. In the backbone, CSPDarknet53 undertakes feature extraction on the image, while the neck incorporates SPP, executing max pooling with four different pool sizes ($5\times5$, $9\times9, 13\times13$, $1\times1$)[38]. Post-SPP, another neck component, PANet, enhances the feature map's quality by performing upsampling and downsampling using
both low-level and high-level feature maps. The head section executes the object detection process across three different scales: the first YOLO head for small objects, the second for medium-sized objects, and the third for large objects[40]. The architecture of YOLOv4 is illustrated in Figure 3.

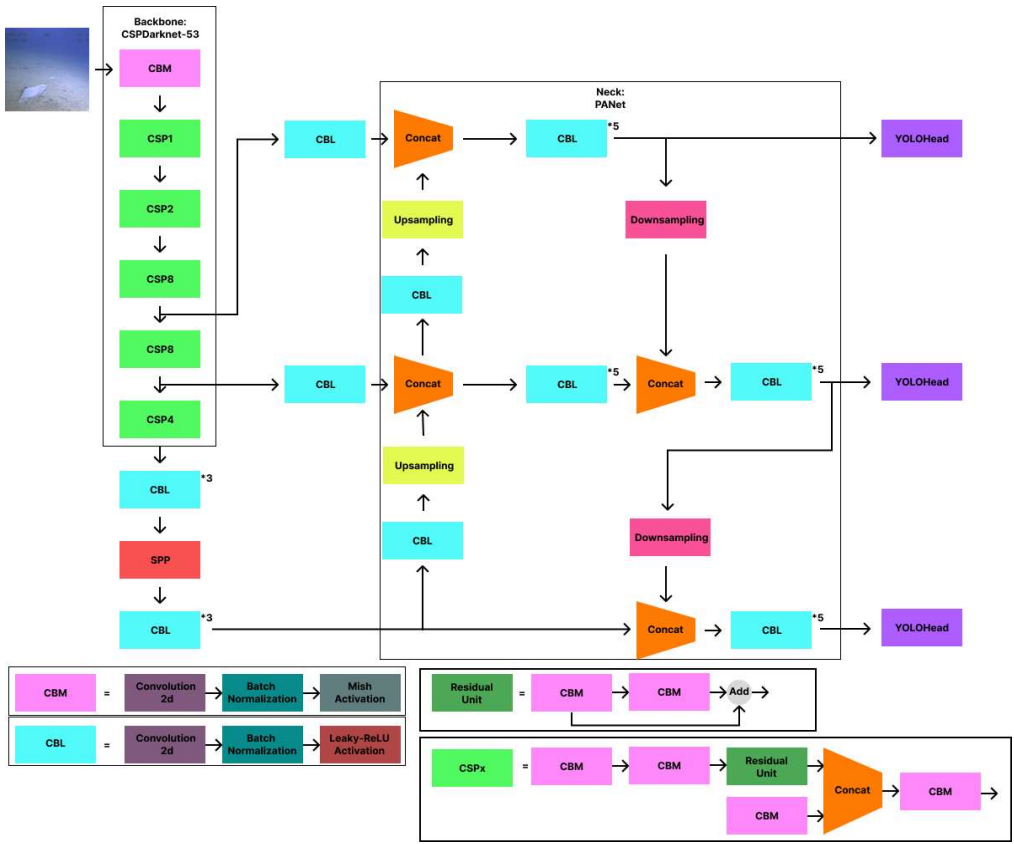

Figure 3: YOLOv4 Architecture

## 3.2. Loss Function

In the architecture of YOLOv4, the selection of a loss function plays a pivotal role in optimizing the model's performance. YOLOv4 incorporates four types of loss functions to efficiently train the network. These encompass the CIoU loss [41], object confidence loss, no-object confidence loss, and classification loss [40], each addressing distinct aspects of the detection process.

The CIoU loss, or Complete Intersection over Union loss, is a distinctive addition in YOLOv4 and is employed to refine the calculation of Intersection over Union (IoU) scores. The formula for CIoU loss includes the integration of specific parameters such as the coordinates (x, y) and the dimensions (width and height) of bounding boxes. This refinement in the loss formula signifies a nuanced approach to assessing the agreement between predicted and ground truth bounding boxes, thereby contributing to more accurate and precise object localization[41].

$$L_{CIoU}(\text{Complete IoU loss}) = 1 - IoU + \frac{\rho(b, b^{gt})}{c^2} + \alpha v$$

(1)

Where ($\rho$) denotes the Euclidian distance. ($b$) and ($b^g t$) denote the central point of the predicted box and ground truth box. $c$ is the length of the shortest enclosing box that covering the two boxes. ($\alpha$) is a positive trade-off parameter, and ($v$) measures the consistency of aspect ratio. The equation ($\alpha$) can be defined as

$$\alpha(\text{positive tradeoff parameter}) = \frac{v}{(1 - IoU) + v}$$

(2)

The equation $v$ can be defined as

$$v(\text{consistency of aspect ratio}) = \frac{4}{\pi^2}\left(\arctan \frac{w^{gt}}{h^{gt}} - \arctan \frac{w}{h}\right)^2$$

(3)

Where $w$ denotes the width, $h$ denotes the height, and $g_t$ denotes the ground truth. The object confidence loss and no object confidence loss use binary cross entropy that calculates the confidence prediction and confidence ground truth. Confidence ground truth becomes 1 if in that box there is an object but becomes 0 if in that box there is no object. The equation of object confidence loss can be defined as

$$Obj = \sum_{i=0}^{s^2} \sum_{j=0}^{B} I_{ij}^{obj}[\hat{C}_i \log C_i + (1 - \hat{C}_i)\log(1 - C_i)]$$

(4)

Where $s^2$ is defined as the total number of grid cells, and since the image's aspect ratio is 1:1, the formulas is $s \times s$ or $s^2$. B is defined as the total number of bounding boxes. i is defined as specific cell location of the grid, j is defined as the bounding box accessed within the cell. $I_{ij}^{noobj}$ is a function, defined as 1 if the cell i inside the j bounding box contain objects, and 0 otherwise. The equation of no-object confidence loss can be defined as

$$\text{Noobj} = \sum_{i=0}^{s^2} \sum_{j=0}^{B} I_{ij}^{noobj} \left[ \hat{C}_i \log C_i + (1 - \hat{C}_i) \log (1 - C_i) \right]$$

(5)

Where the variable is remaining the same except the $I_{ij}^{noobj}$ defined as 1 if the cell i inside the j bounding box is not contain objects, and 0 otherwise. $\hat{C}_i$ is defined as predicted confidence, $C_i$ is defined as actual confidence Classification loss uses binary cross entropy to calculate the class prediction that has implemented one-hot encoding. The equation of classification can be defined as

$$\text{class} = \sum_{i=0}^{s^2} I_{ij}^{obj} \sum_{c \in \text{classes}} [\hat{p}_i \log p_i + (1 - \hat{p}_i) \log (1 - p_i)]$$

(6)

$\hat{p}_i$ is defined as predicted classification, $p_i$ is defined as actual classification.

### 3.3. Channel Pruning

After several years of deep learning development, encompassing architectures such as AlexNet[42], VGG[43], GoogLeNet[44], U-Net[45], MaskRCNN[46], and others, the architecture of neural networks has grown larger. This growth has implications for the computational load, making it increasingly challenging to deploy these architectures on low-level hardware. Pruning emerges as a solution to manage this significant computational load and reduce it. Pruning can be applied at various levels, commencing with weights and extending to layers[47]. It is evident that pruning at the most granular level, such as weights, will yield the most flexible pruning. Nevertheless, it is essential to ensure adequate preparation, including the provision of software and hardware accelerators. At larger levels, such as layers, pruning is less flexible but is the most straightforward to perform. Channel pruning, as one level of pruning, focuses on reducing a model's size by pruning its channels. In recent years, a diverse array of models has been developed, featuring an ever-increasing number of parameters and convolutions, which translates to a substantial computational cost for training. Pruning aims to eliminate unimportant weights, thus reducing the overall computational burden.

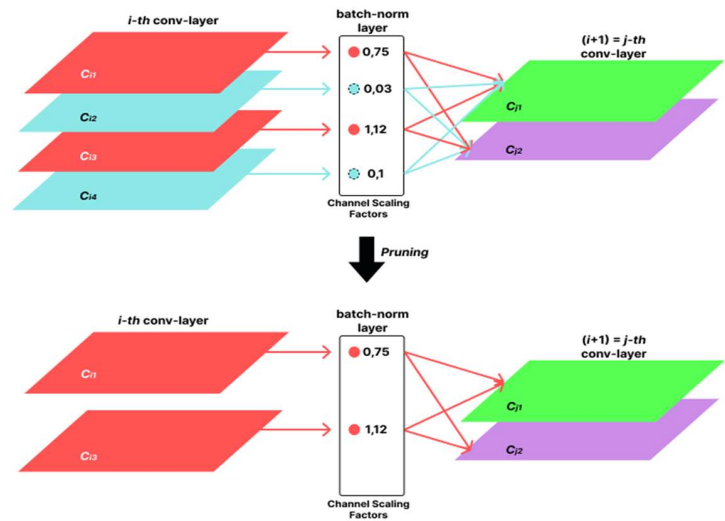

Figure 4: Channel Pruning Diagram

Channel pruning removing channels that are considered less crucial. The importance of each channel can be determined through sparse regularization, which identifies channels that are infrequently utilized. This is achieved by inserting L1-regularization into batch normalization, enabling the batch normalization layer to store the scaling factor that determines the usage and level of importance of the channel [48]. Channel with minimal importance will have a value that is close to zero. After training process that included sparse regularization ends, channel pruning is performed by pruning the layer with a scaling factor close to zero, and pruning the channel's layers as shown in Figure 4. It is interesting that the accuracy result of the channel pruning is still amazing and importantly the parameters decreased up to 20x from the base model. Since it is using several models such as VGGNet, DenseNet, and ResNet-164, the writer is interested about applying channel pruning inside YOLOv4

## 4. Experiments

The research conducted training using six distinct schemes, each designed to explore and optimize the YOLOv4 architecture for marine debris detection. The breakdown of these schemes is as follows:

・ Scheme 1: YOLOv4 from Scratch

Training YOLOv4 model without leveraging pre-existing weights or knowledge, starting the learning process from scratch.

- Scheme 2: Pretrained YOLOv4

    Utilizing pre-existing weights and knowledge by training YOLOv4 on a pretrained model, enhancing the model's ability to generalize and learn patterns effectively.

- Scheme 3: YOLOv4 with freezing backbone layer

    Training YOLOv4 with a frozen backbone layer, allows the model to fine-tune specific aspects without altering the foundational features learned during the initial training.

- Scheme 4: YOLOv4 with freezing backbone+neck layer

    Extending the freezing approach to both the backbone and neck layers, aiming to selectively refine higher-level features in the model while preserving lower-level representations.

- Scheme 5: YOLOv4-tiny from Scratch

    Employing a smaller variant, YOLOv4-tiny, and training it from scratch, exploring the trade-off between model complexity and detection performance.

- Scheme 6: Pruned-YOLOv4

    Implementing channel pruning techniques on YOLOv4 to create a pruned version, seeking to enhance efficiency without significantly compromising detection accuracy.

Each scheme represents a unique approach to training the YOLOv4 architecture, allowing for a comprehensive analysis of different strategies and their impact on marine debris detection performance.

### 4.1. Scheme 1

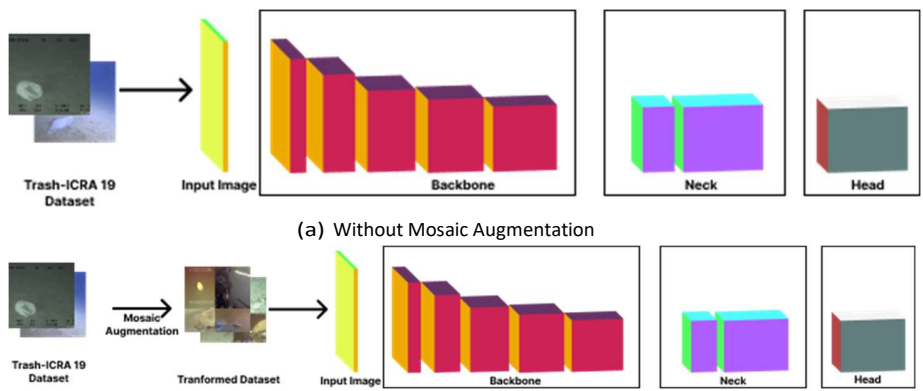

(a) Without Mosaic Augmentation

(b) With Mosaic Augmentation

Figure 5: Schema 1 of Training YOLOv4 from Scratch

The model employed in Scheme 1 is YOLOv4 from Scratch. This scheme also includes a comparison with and without mosaic augmentation. Scheme 1 is depicted in Figure 5. After training for 1000 epochs, the loss values can be observed in Figure 6.

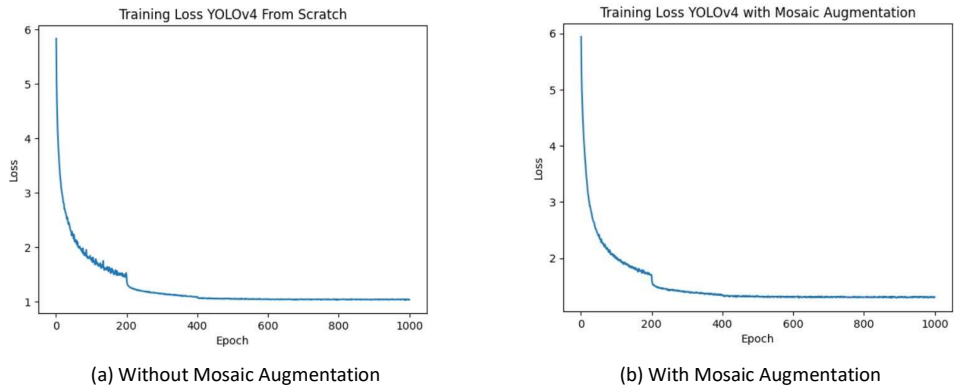

(a) Without Mosaic Augmentation    (b) With Mosaic Augmentation

Figure 6: Loss value on Schema 1

The loss values for both training sets decrease rapidly until epoch 200, after which the reduction in loss occurs more gradually. The model converges after epoch 400. The sudden drop in loss at epoch 201 is attributed to the adjustment of the learning rate value every 200 epochs.

| Without Mosaic Augmentation | | | With Mosaic Augmentation | | |
| --- | --- | --- | --- | --- | --- |
| AP per Class | mAP | FPS | AP per Class | mAP | FPS |
| AP Plastic: 0.981<br>AP Bio: 0.965<br>AP ROV: 0.959 | 0.9684 | 15.63 | AP Plastic: 0.9871<br>AP Bio: 0.9704<br>AP ROV: 0.9463 | 0.968 | 15.32 |

Table 1: Schema 1 Training Result

In Table 1, it is evident that the mAP value for Scheme 1 without mosaic augmentation reached 96.84% with an FPS value of 15.63. The mAP for each class also yielded high values above 95%, with the highest AP in the plastic class reaching 98.1%. The lowest AP is observed in the ROV class with a value of 95.93%.

For Scheme 1 with mosaic augmentation, the mAP value reached 96.6% with an FPS value of 15.32. The mAP for each class also demonstrated high values above 95%, with the highest AP in the plastic class reaching 98.7%. The lowest AP is in the ROV class with a value of 94.63

*4.2. Scheme 2*

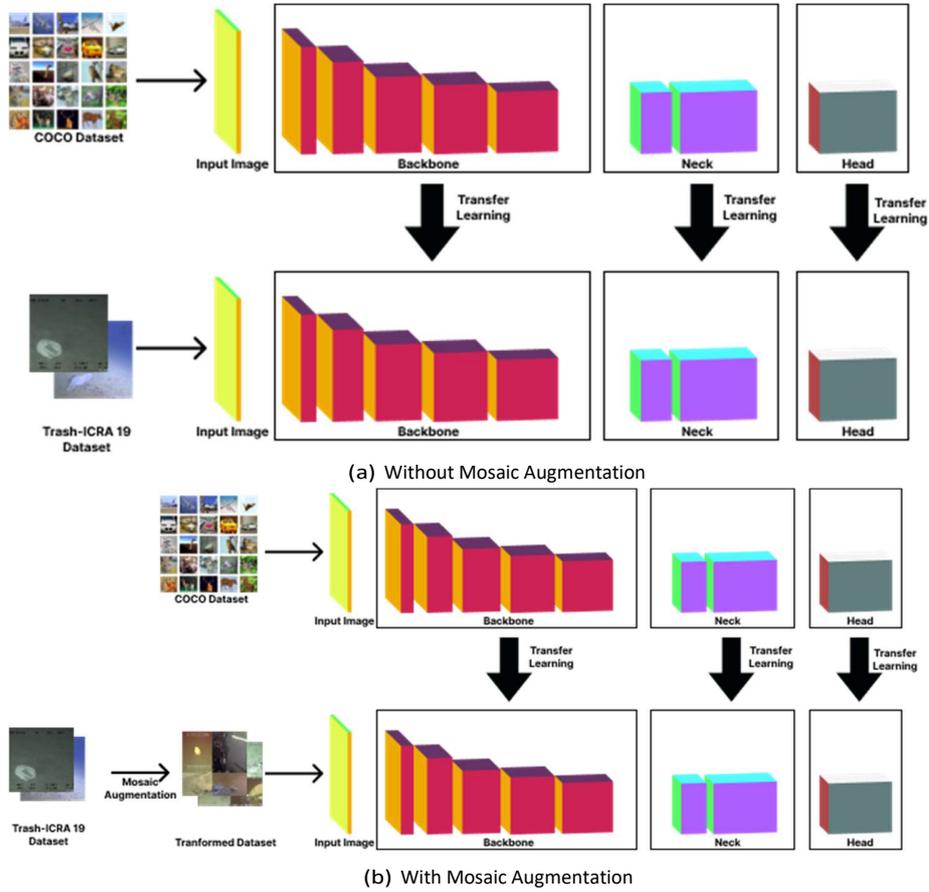

Figure 7: Schema 2 of fine-tuning pretrained YOLOv4

Scheme 2 is a model training scheme using YOLOv4 with pre-training from the COCO[50] dataset. In Scheme 2, a comparison of the pre-trained YOLOv4 model without mosaic augmentation and with mosaic augmentation is conducted. The training progress for each step is illustrated in Figure 7. The corresponding loss values are depicted in Figure 8, and the mAP results are presented in Table 2.

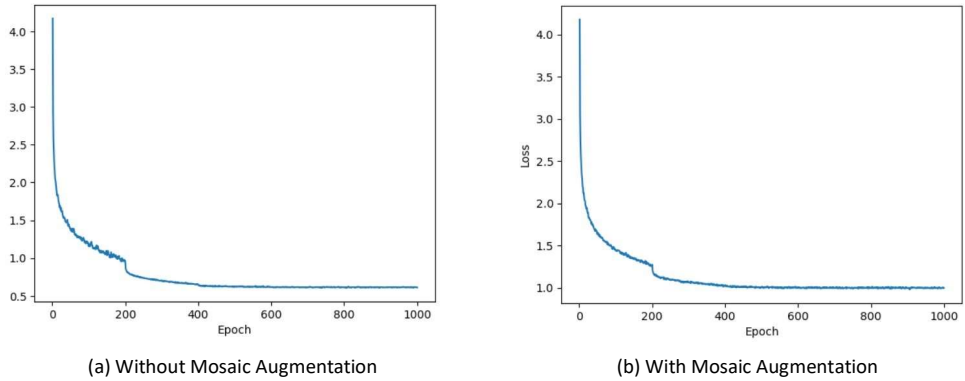

| (a) Without Mosaic Augmentation | (b) With Mosaic Augmentation |

Figure 8: Loss Value of YOLOv4 Pretrained

The Figure 8 shows how the schema 2 loss value is moving the same as how the schema 1 moving

| Without Mosaic Augmentation | | | With Mosaic Augmentation | | |
|---|---|---|---|---|---|
| AP per Class | mAP | FPS | AP per Class | mAP | FPS |
| AP Plastic: 0.988<br>AP Bio: 0.9762<br>AP ROV: 0.9638 | 0.976 | 15.94 | AP Plastic: 0.9884<br>AP Bio: 0.9813<br>AP ROV: 0.9538 | 0.9745 | 15.19 |

Table 2: Schema 2 Training Result

Table 2 shows that the highest mAP value for training without Mosaic Augmentation was 97.6% with FPS value of 15.94. The mAP of each class also produced high values above 95%, with the highest AP in the plastic class reaching 98.8%. The lowest AP is in the ROV class with a value of 96.38%.

The mAP value for schema 2 with mosaic augmentation reached 97.45% with FPS value of 15.19. The mAP of each class also produced high values above 95%, with the highest AP in the plastic class reaching 98.84%. The lowest AP is in the ROV class with a value of 95.38%.

*4.3. Scheme 3*

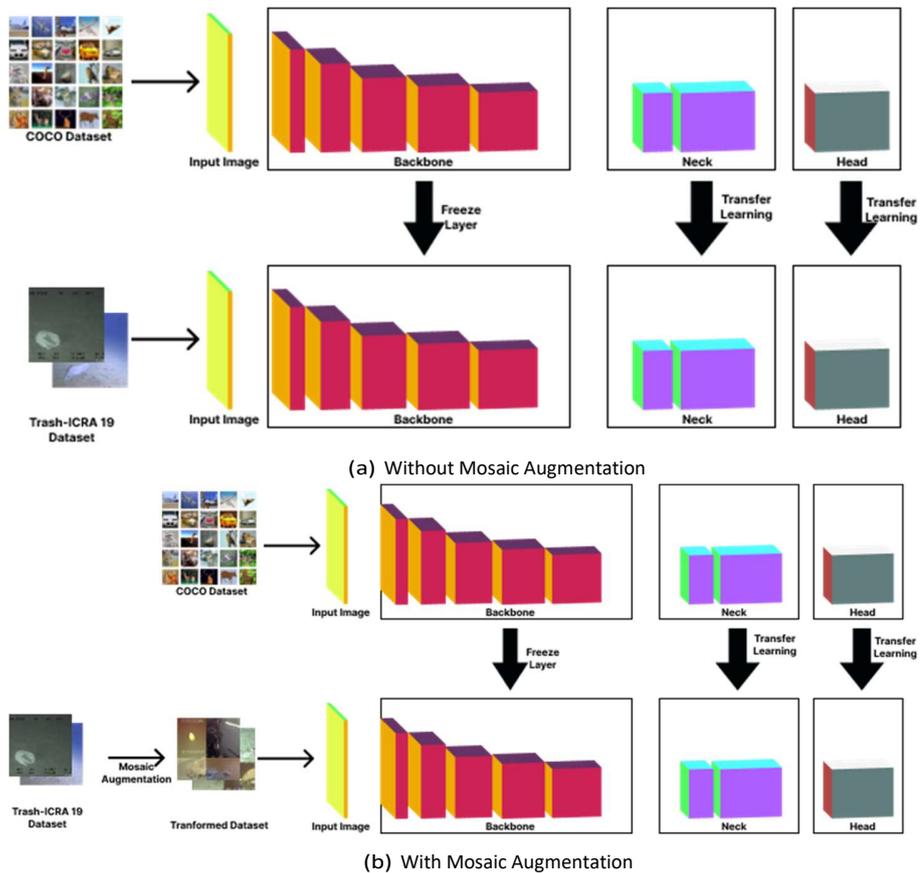

Figure 9: Schema 3 of fine-tuning YOLOv4 with frozen backbone layer

Scheme 3 is a model training scheme that is similar to scheme 2, but it includes a frozen layer on the backbone only. The step of each training can be seen at Figure 9. The loss values are described in Figure 10, with the mAP results described in Table 3

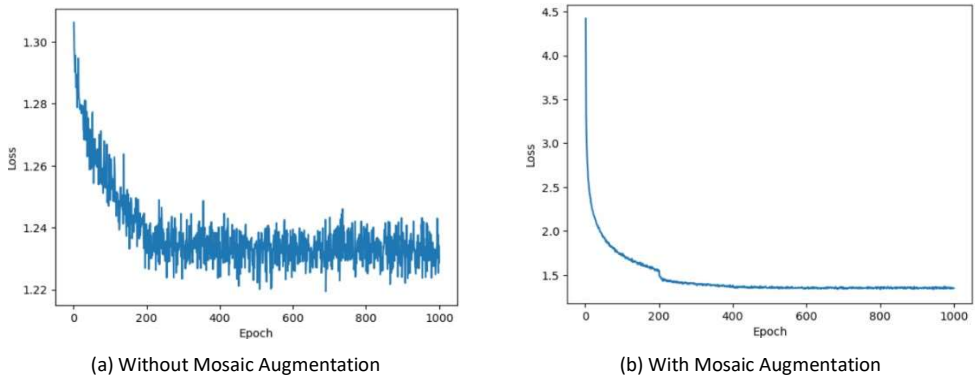

(a) Without Mosaic Augmentation  (b) With Mosaic Augmentation

Figure 10: Loss Value of YOLOv4 Freeze Backbone Layer

Figure 10 depicts a notable decrease in the loss value during the initial 200 epochs of the two training processes. However, at epoch 201, a sharp decline occurred, after which the decrease stabilized until epoch 400.

| Without Mosaic Augmentation | | | With Mosaic Augmentation | | |
| --- | --- | --- | --- | --- | --- |
| AP per Class | mAP | FPS | AP per Class | mAP | FPS |
| AP Plastic: 0.9815　AP Bio: 0.974　AP ROV: 0.9335 | 0.963 | 14.09 | AP Plastic: 0.9858　AP Bio: 0.976　AP ROV: 0.934 | 0.9652 | 14.88 |

Table 3: Schema 3 Training Result

Table 3 shows that the highest mean average precision (mAP) value for training without Mosaic Augmentation was 96.3% with FPS value of 14.09. The mAP of each class also produced high values above 90%, with the highest AP in the plastic class reaching 98.15%. The lowest AP is in the ROV class with a value of 93.35%.

The mAP value for schema 3 with mosaic augmentation reached 96.52% with FPS value of 14.88. The mAP of each class also produced high values above 90%, with the highest AP in the plastic class reaching 98.58%. The lowest AP is in the ROV class with a value of 93.5%.

*4.4. Scheme 4*

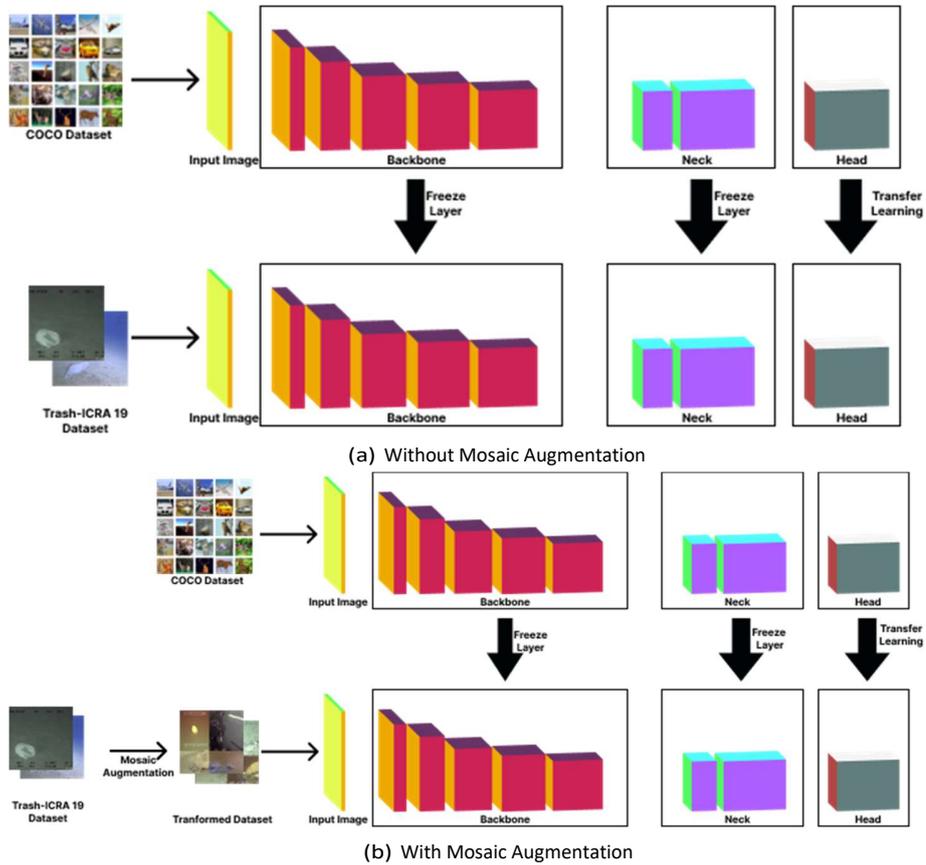

Figure 11: Schema 4 of YOLOv4 with frozen backbone+neck layer

Scheme 4 is a model training approach similar to scheme 3, utilizing a frozen layer, but it adds the frozen layer to the neck. The steps of each training iteration can be observed in Figure 11. The corresponding loss values are depicted in Figure 12, and the mAP results are detailed in Table 3.

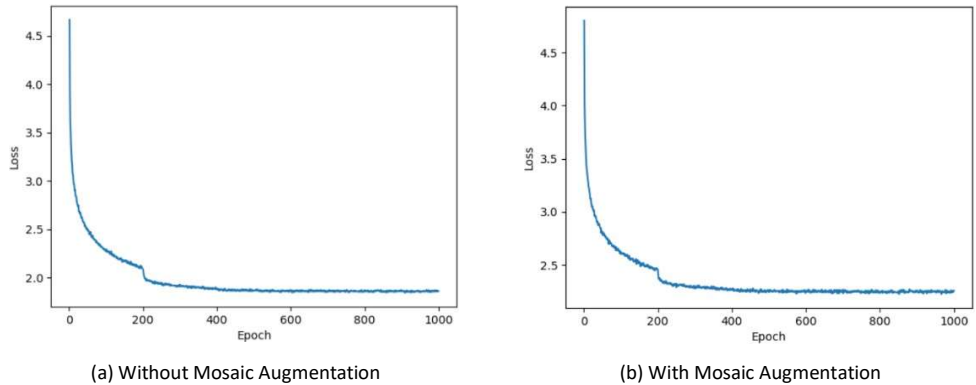

(a) Without Mosaic Augmentation  (b) With Mosaic Augmentation

Figure 12: Loss Value of YOLOv4 Freeze Backbone and Neck Layer

Figure 12 shows a significant decrease in the loss value during the first 200 epochs of both training. However, at epoch 201, there was a sharp decline, after which the decline stabilized until epoch 400.

| Without Mosaic Augmentation | | | With Mosaic Augmentation | | |
| --- | --- | --- | --- | --- | --- |
| AP per Class | mAP | FPS | AP per Class | mAP | FPS |
| AP Plastic: 0.8188<br>AP Bio: 0.873<br>AP ROV: 0.855 | 0.8489 | 14.45 | AP Plastic: 0.8649<br>AP Bio: 0.8647<br>AP ROV: 0.8993 | 0.8763 | 12.99 |

Table 4: Schema 4 Training Result

Table 4 indicates that the highest mean average precision (mAP) value for training without Mosaic Augmentation was 84.9% with an FPS value of 14.45. The mAP of each class yielded lower values than the previous scheme, with the highest AP in the BIO class reaching 87.3%. The lowest AP is in the plastic class with a value of only 81.88%.

The mAP value for schema 4 with mosaic augmentation reached 87.63% with a FPS value of 12.99. The mAP of each class also produced lower values than the previous scheme, with the highest AP in the ROV class reaching 89.93%. The lowest AP is in the Bio class with a value of 86.47%.

*4.5. Scheme 5*

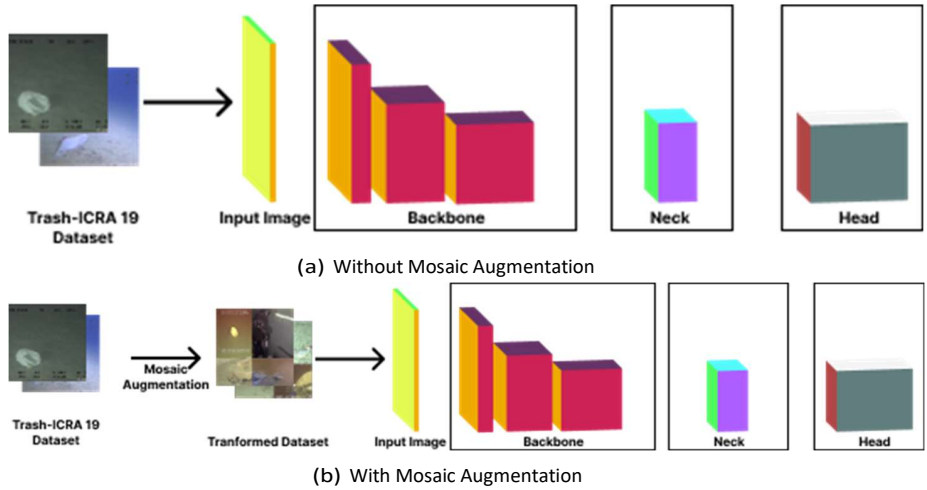

(a) Without Mosaic Augmentation

(b) With Mosaic Augmentation

Figure 13: Schema 5 of training YOLOv4-tiny from scratch

Scheme 5 is a model trained using YOLOv4-tiny from Scratch. It compares the YOLOv4-tiny model with and without mosaic augmentation. The step of each training can be seen at Figure 13. Figure 14 describes the loss values, and Table 5 shows the mAP results.

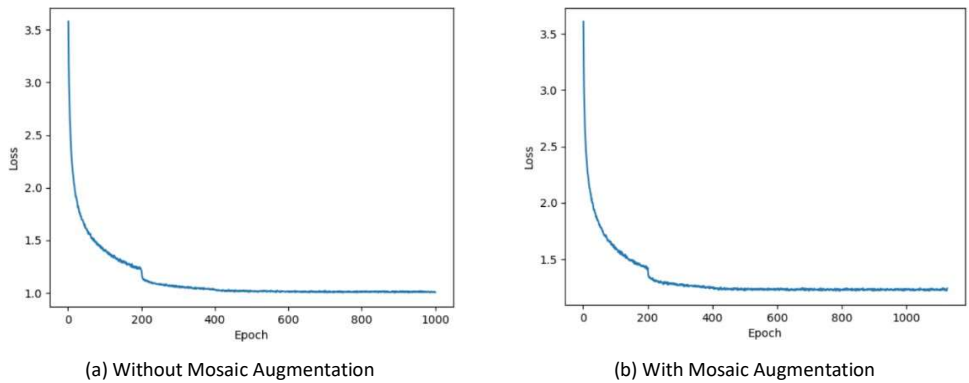

(a) Without Mosaic Augmentation

(b) With Mosaic Augmentation

Figure 14: Loss Value of YOLOv4-tiny

Figure 14 shows a significant decrease in the loss value during the first 200 epochs of the 2 training processes. At epoch 201, the loss value drops abruptly and then stabilizes until epoch 400.

| Without Mosaic Augmentation | | | With Mosaic Augmentation | | |
|---|---|---|---|---|---|
| AP per Class | mAP | FPS | AP per Class | mAP | FPS |
| AP Plastic: 0.963<br>AP Bio: 0.9376<br>AP ROV: 0.9335 | mAP: 0.9447 | 29.22 | AP Plastic: 0.9636<br>AP Bio: 0.9203<br>AP ROV: 0.918 | 0.9339 | 31.65 |

Table 5: Schema 5 Training Result

Table 5 shows that the highest mean average precision (mAP) value for training without Mosaic Augmentation was 94.47% with FPS value of 29.22. The mAP of each class produced high value more than 90%, with the highest AP in the plastic class reaching 96.3%. The lowest AP is in the plastic class with a value of only 93.35%.

The mAP value for schema 5 with mosaic augmentation reached 93.39% with FPS value of 31.65. The mAP of each class produced high value more than 90%, with the highest AP in the plastic class reaching 96.36%. The lowest AP is in the Bio class with a value of 91.8%.

*4.6. Scheme 6*

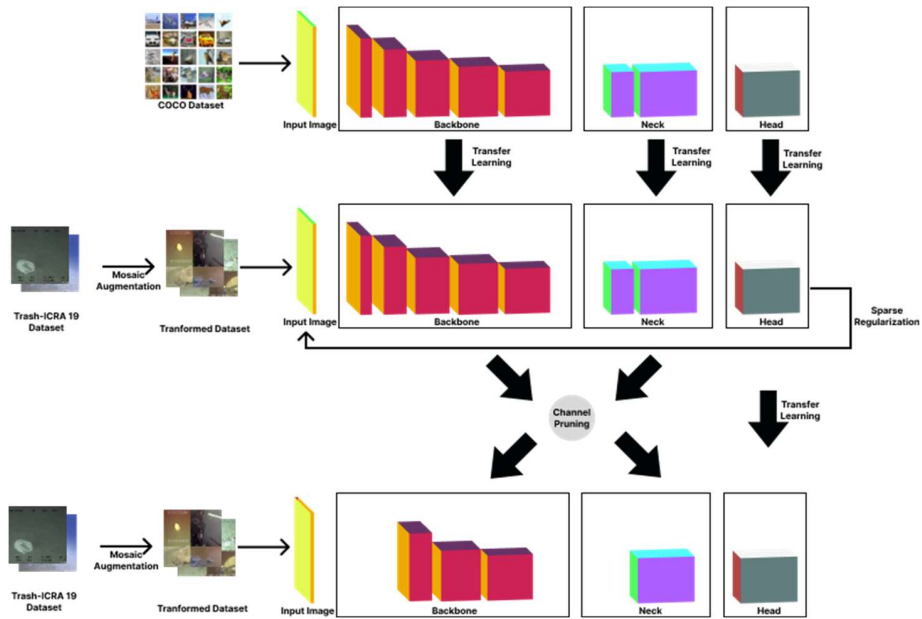

Figure 15: Schema 6 of fine-tuning pruned-YOLOv4

Scheme 6 involves channel pruning of the fine-tuned model in scheme 2 with mosaic augmentation. The model is pruned by 20% and 50%. The step of each training can be seen at Figure 15. Figure 16 shows the corresponding loss values, while Table 6 presents the mAP results.

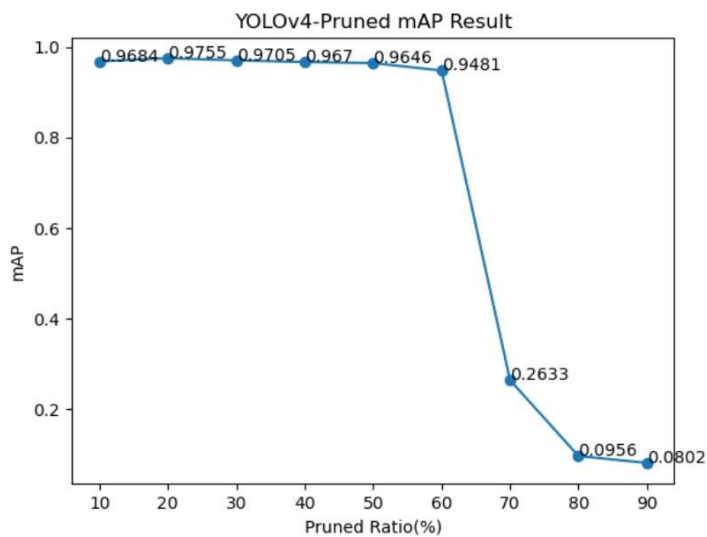

Figure 16: YOLOv4-Pruned mAP Result Graph

Figure 16 shows the comparison of mAP values with the pruning ratio. The highest mAP value is achieved at a pruning ratio of 20%. After that, the mAP value decreases until it experiences a significant drop at a pruning ratio of 80%.

| Pruning Ratio | mAP | FPS | Parameter |
|---|---|---|---|
| 10% | 96.84% | 15.7 | 52.23 M |
| <u>20%</u> | <u>97.55%</u> | <u>16.01</u> | <u>44.64 M</u> |
| 30% | 97% | 16.8 | 32.48 M |
| 40% | 96.7% | 18.2 | 24.39 M |
| **50%** | **96.46%** | **19.4** | **17.66 M** |
| 60% | 94.81% | 21.2 | 11.93 M |
| 70% | 26.33% | 12.2 | 6.98 M |
| 80% | 9.56% | 27.9 | 3.16 M |
| 90% | 8.02% | 29.4 | 0.76 M |

Table 6: Schema 6 Training Result. Bold result denotes the most efficient among all the pruned model. Underline result denotes the highest mAP among all the pruned model

Table 6 illustrates that the mAP value decreases with each increase in pruning ratio, particularly at 70% pruning. The highest mAP value of 97.55% was achieved at 20% pruning, while the lowest mAP value of only 8.02% was obtained at 90% pruning.

Additionally, the FPS value increased with an increasing pruning ratio, except at 70% pruning. The decrease in FPS observed at 70% pruning was caused by the model predicting an excessive number of bounding boxes, as indicated by the low mAP value. This led to a larger computational load required to generate these bounding boxes. Furthermore, at 80% and 90% pruning, the model produced fewer bounding boxes, resulting in a higher FPS.

Among all the pruning ratios, the 20% and 50% pruning ratios are superior to the others. The highest mAP was achieved with a 20% pruning, while 50% resulted in better FPS with only a slight decrease in mAP.

*4.7. Analysis*

Table 7 presents a model comparison based on the best mAP and FPS obtained by testing all schemes 1-5. The highest mAP value, 97.6%, was achieved when testing with YOLOv4 Pretrained. Interestingly, the use of mosaic augmentation resulted in a decrease in mAP values for most models, with YOLOv4 from Scratch and YOLOv4 Pretrained experiencing a decrease of around 0.2%, and YOLOv4 Tiny experiencing a decrease of 1%. Notably, only the YOLOv4 freeze layer showed an increase in mAP. The mAP increase in YOLOv4 Pretrained Freeze Backbone+Neck reached 2.8%, while in YOLOv4 Pretrained Freeze Backbone, it only reached 0.2%.

Based on mAP, the best results are achieved by using pretrained models and then fine-tuning them. The highest mAP is obtained both with and without mosaic augmentation. The model with the lowest mAP is YOLOv4 pretrained freeze backbone+neck, as weight updates are only performed on the head layer.

In terms of FPS, YOLOv4 achieves the best FPS with YOLOv4 Tiny with Mosaic Augmentation, with a value of 31.65. The Backbone+Neck model had the lowest FPS. The number of parameters or complexity of the model also affects the FPS value. When comparing YOLOv4 with YOLOv4 Freeze Backbone+Neck, the model's mAP value is directly proportional to its FPS. When comparing YOLOv4-tiny with YOLOv4, the number of parameters is inversely proportional to FPS.

| Model | Without Mosaic Augmentation | | With Mosaic Augmentation | |
|---|---|---|---|---|
| | Best mAP | FPS | Best mAP | FPS |
| YOLOv4 from Scratch | 96.84% | 15.63 | 96.8% | 15.32 |
| Pretrained YOLOv4 | **97.6%** | 15.94 | 97.45% | 15.19 |
| Pretrained YOLOv4+ Frozen Backbone | 96.3% | 14.09 | 96.5% | 14.88 |
| Pretrained YOLOv4+ Frozen Backbone+Neck | 84.89% | 14.45 | 87.63% | 12.99 |
| YOLOv4-tiny from Scratch | 94.47% | 29.22 | 93.39% | **31.65** |

Table 7: Scheme 1-5 mAP and FPS Training Result Summary

In Scheme 6, the research was conducted after Scheme 2 was completed. The results of channel pruning can be found in Table 8.

| Model | mAP | FPS | Parameter |
|---|---|---|---|
| Pretrained YOLOv4 with Mosaic Aug | 97.4% | 15.19 | 63.95 M |
| YOLOv4 Tiny | 94.4% | 31.65 | 9.34 M |
| YOLOv4 20% Pruning | 97.5% | 16.01 | 44.64 M |
| YOLOv4 50% Pruning | 96.4% | 19.4 | 17.66 M |

Table 8: Scheme 6 Training Result Summary

After pruning, the best mAP is achieved by YOLOv4 with 20% pruning, reaching a value of 97.5%, which is a 0.1% increase compared to the base model.

Channel pruning contributes to an increase in the FPS value, with YOLOv4 at 50% pruning achieving an FPS of 19.4 compared to the base model's 15.19 FPS. YOLOv4-tiny, with fewer parameters exhibits better FPS.

The comparison of Average Precision (AP) per class for all schemes can be found in Table 9.

| Model | Without Mosaic Augmentation | | | With Mosaic Augmentation | | |
|---|---|---|---|---|---|---|
| | Plastic AP | Bio AP | ROV AP | Plastic AP | Bio AP | ROV AP |
| YOLOv4 from Scratch | 98.1% | 96.5% | 95.9% | 98.71% | 97.04% | 94.63% |
| YOLOv4 Pretrained | 98.8% | 97.62% | **96.38%** | 98.84% | **98.13%** | 95.38% |
| YOLOv4 Pretrained Freeze Backbone | 98.15% | 97.4% | 93.35% | 98.58% | 97.6% | 93.4% |
| YOLOv4 Pretrained Freeze Backbone+ Neck | 81.88% | 87.3% | 85.5% | 86.49% | 86.47% | 89.93% |
| YOLOv4 Tiny | 96.3% | 93.76% | 93.35% | 96.36% | 92.03% | 91.8% |
| YOLOv4 20% Pruning | - | - | - | **99%** | 97.72% | 95.91% |
| YOLOv4 50% Pruning | - | - | - | 98.6% | 95.6% | 95.2% |

Table 9: Scheme 1-6 AP per Class Training Result Summary

Table 9 reveals that the highest Average Precision (AP) is achieved in the plastic class. Generally, most models yield the highest AP in the plastic class compared to other classes, except for models with a freeze layer. The YOLOv4 model with 20% pruning attains the best result in the plastic class with 99%. Meanwhile, the YOLOv4 Pretrained model with mosaic augmentation obtains the highest AP in the Bio class at 98.13%, and the YOLOv4 Pretrained model without mosaic augmentation performs best in the ROV class with 96.38

The lowest AP values across all classes are observed in the YOLOv4 Pretrained Freeze Backbone + Neck model. The plastic and ROV classes exhibit the lowest AP when the model does not use mosaic augmentation, while the Bio class shows the lowest AP when the model uses mosaic augmentation.

| Research (GPU) (Dataset) | Model | mAP | FPS |
|---|---|---|---|
| Robotic Detection of Marine Litter Using Deep Visual Detection Models[28] (TX2) (Trash ICRA-19) | YOLOv2, Tiny-YOLO, Faster R-CNN, SSD | 47.9% 31.6% 81% 67.4% | 6.2 20.5 5.66 11.25 |
| Research on underwater object recognition based on YOLOv3[20] (GTX Titan X) (URPC (Underwater Robot Picking Contest)) | Faster R-CNN, YOLOv3 | 69.7% 76.1% | 8 20 |
| Pruning-Based YOLOv4 Algorithm for Underwater Garbage Detection[31] (GTX 1080 Ti) (Underwater Image Dataset (private)) | Pruned-YOLOv4, YOLOv4, YOLOv3 | 90.3% 91.3% 89.7% | 58.8 43.3 38 |
| Proposed Research (GTX 1050 Mobile) (Trash ICRA-19) | YOLOv4, YOLOv4-tiny, YOLOv4 20% Pruning, YOLOv4 50% Pruning | 97.6% 94.4% 97.5% 96.4% | 15.19 31.65 16.01 19.4 |

Table 10: Comparison with Previous Research

Table 10 presents a comparison between previous research and the proposed research where the proposed research show improvements in several areas proved by comparison in mAP, FPS, and the GPU used when compared to previous research.

In comparison with Fulton et al.'s study (2019)[28] that used the same dataset, since the SSD overcome mAP result than YOLOv2 and Tiny-YOLO model, the proposed model, YOLOv4 50% Pruning successfully overcome the mAP and increasing it from 67.4% to 96.4%. Compared to the FPS values between this two research, the proposed research has competitive result with tiny-YOLO from Fulton et al.'s study which is 19.4 FPS and 20.5FPS. Since the proposed model use testing with GTX 1050 mobile, Fulton et al.'s study uses Jetson TX2 which is compatible with robots and object detection. This result shows how the proposed model can still achieve real-time detection speeds proven by lower gap on proposed method but only 1.1FPS difference with the tiny-YOLO result in Fulton's study.

Similarly, the proposed model(YOLOv4 50% Pruning) in our research yields a higher mAP value compared to the YOLOv3 model used in same domain topic dataset research by Yang et al. (2021)[20] between 96.4% and 76.1%. It has a competitive result also by the FPS because of the GPU power difference where the proposed model has small result but only 0.8FPS difference with Yang et al. study(2021)[20] that used GTX Titan X. It shows that YOLOv4 with low GPU technology that causing a lower FPS than its original performance, still can have the competitive result with YOLOv3 if implemented with pruning method and having mAP result that remain the same as base YOLOv4

Furthermore, when comparing with Tian et al., 2021[31] research who also implementing pruning in YOLOv4 and used same domain topic dataset but with better GPU, has a significant difference FPS between 43.3FPS and 15.19FPS from the base model used in proposed. It is also shows from both research that adding pruning successfully increase the FPS score with a small mAP reduction, and it is also shows that YOLOv4 50% Pruning is still applicable for real-time detection even with a low GPU power.

Figure 17 displays some detection results of all scheme models. Since most of the models have mAP and AP per class above 90%, the images in Figure 17 successfully detect the corresponding objects. The models are less successful in detecting correct objects only in the YOLOv4 freezing backbone+neck model, as the mAP values of these models are below 90%.

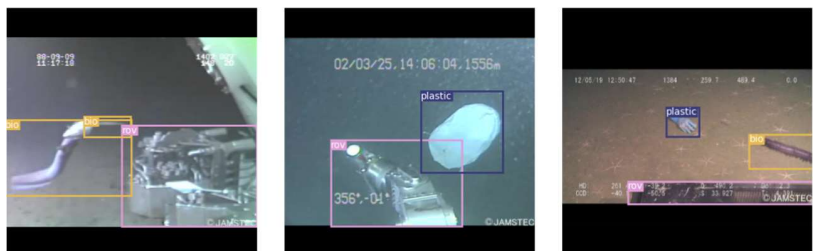
(a) YOLOv4 from Scratch Detection (mAP: 99.99%)

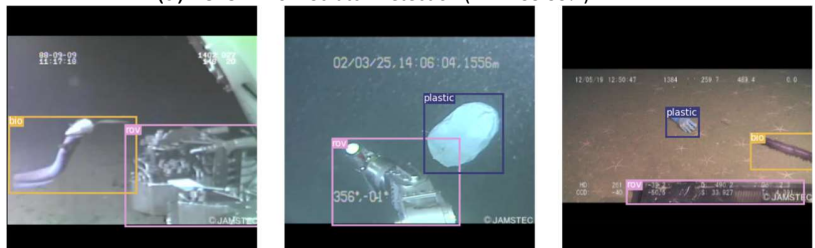
(b) YOLOv4 from Scratch with Mosaic Augmentation Detection (mAP: 99.99%)

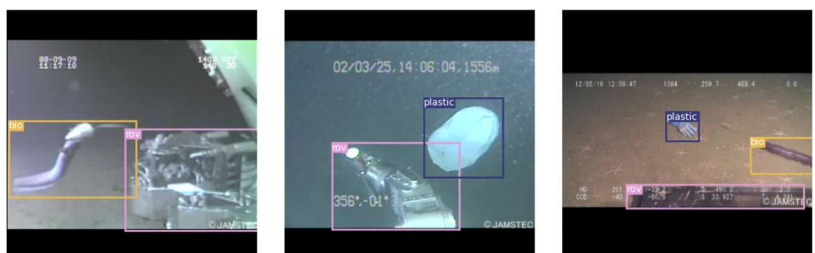
(c) Pretrained YOLOv4 Detection (mAP: 99.99%)

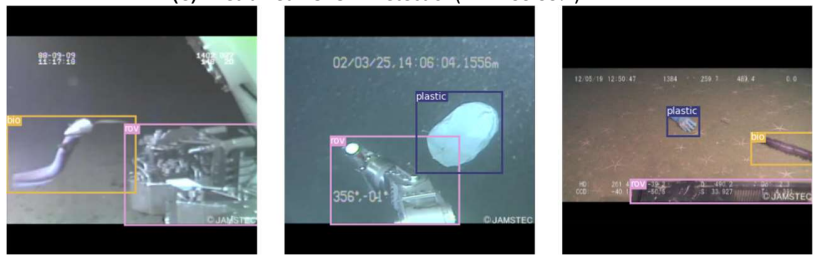

**(d)** Pretrained YOLOv4 with Mosaic Augmentation
Detection (mAP: 99.99%)

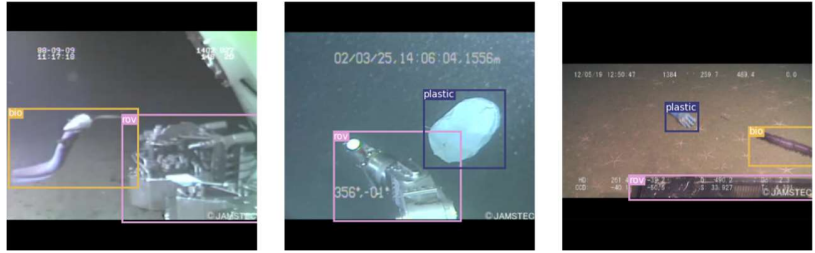

**(e)** Pretrained YOLOv4 with frozen Backbone Layer
Detection (mAP: 99.99%)

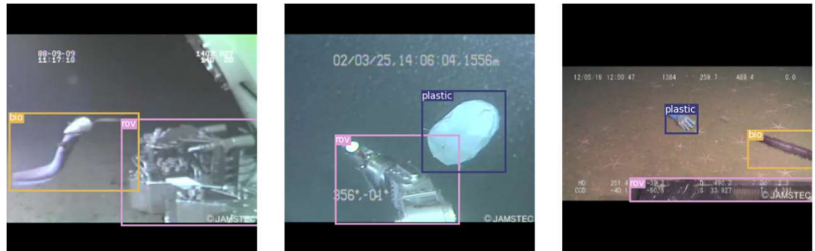

**(f)** Pretrained YOLOv4 with frozen Backbone Layer with
Mosaic Augmentation Detection (mAP: 99.99%)

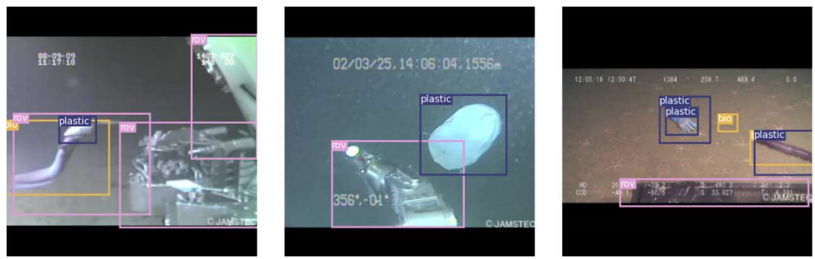

**(g)** Pretrained YOLOv4 with frozen Backbone+Neck Layer
Detection (mAP: 98.61%)

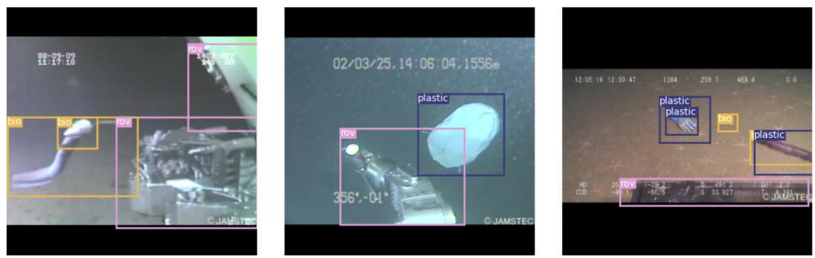

**(h)** Pretrained YOLOv4 with frozen Backbone+Neck Layer
with Mosaic Augmentation Detection (mAP:98.61%)

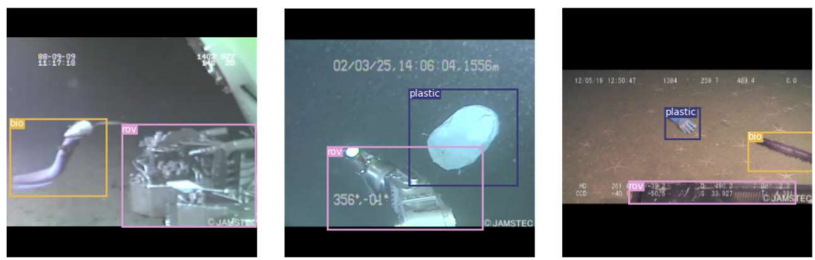

**(i)** YOLOv4-tiny Detection (mAP: 99.99%)

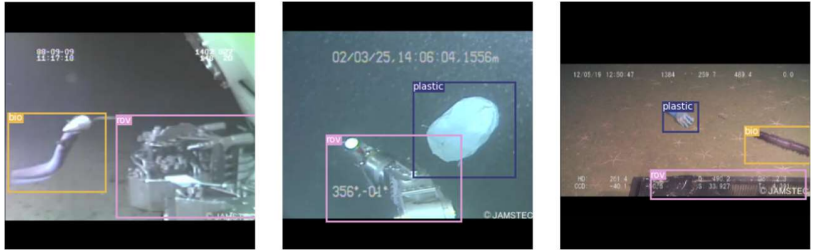

(j) YOLOv4-tiny with Mosaic Augmentation
Detection (mAP: 99.99%)

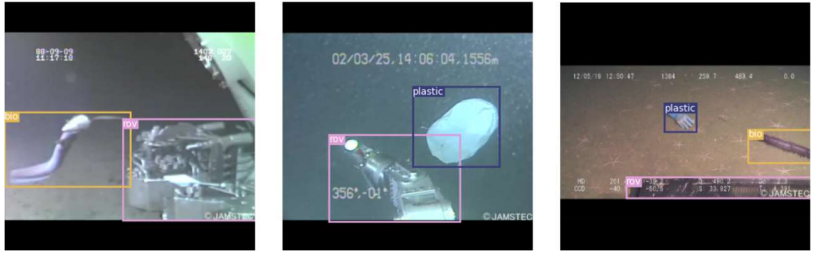

(k) YOLOv4 with 20% Pruning Detection (mAP: 99.99%)

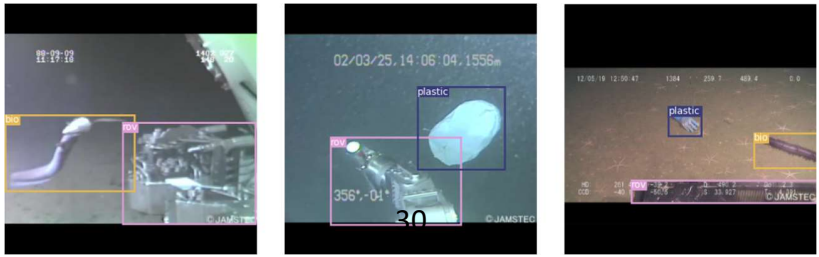

(l) YOLOv4 with 50% Pruning Detection (mAP: 99.99%)

Figure 17: Prediction Result

The mAP value shown in the caption of Figure 17 is the mAP with 3 images shown in Figure 17, where the mAP value reaches 99.99% in almost all models except the YOLOv4 with pretrained freeze backbone+neck layer. However, the value obtained from the model is also very high at 98.61%. The color of the box represents the detected class, where the blue box is the plastic class, pink is the ROV class, and orange is the bio class.

## 5. Conclusion

Based on the test results of the six schemes, it can be concluded that an efficient YOLOv4 model for marine debris object detection can be designed by performing channel pruning on the trained model with the mAP resulting 96.4%. This method increases the FPS of the model while only decreasing the mAP by approximately 1.2% from the base model 97.6% to 96.4%.

The YOLOv4 base model underwent several changes in the six test schemes, including the use of pretraining, freezing layers, and channel pruning. Based on the tests conducted, the model that achieved the highest mAP is pretrained YOLOv4 model with score of 97.6% while for the highest FPS is YOLOv4-tiny with the score 31.65FPS which is developed by focusing for low GPU user. This is show that the proposed model, YOLOv4 50% pruning model proven not the best in each accuracy and speed. However, the proposed model is the most efficiency model because it has an almost similar with the best mAP model with gap 1.2% lower only which is higher than YOLOv4-tiny and offer a notable increase in speed which is higher than pretrained YOLOv4 model.

From the comparation with previous research, it shows that GPU performance is one of the aspects that affect the FPS result. However, since the proposed research use GTX 1050 Mobile that lower than the compared research, the efficient YOLOv4 using 50% pruning was able to successfully overcome the detection object that utilize high GPU performance proved by competitive result with the gap only 1.1FPS and 0.8FPS lower than the compared research. The result also shows that with some improvement, the base YOLOv4 model can be used for real-time detection for marine debris cases even with low GPU, while maintaining a near-identical mAP value as the original YOLOv4

To improve this research in the future, several suggestions can be considered, such as conducting tests by replacing the backbone and neck layers of YOLOv4 with CSPResNext50 or EfficientNet-B3 on the backbone, and FPN or SFAM on the neck to observe variations. Perform pruning experiments with layer pruning in addition to channel pruning.

**References**

[1] M. Z. Elamin, K. N. Ilmi, T. Tahrirah, Y. A. Zarnuzi, Y. C. Suci, D. R. Rahmawati, D. M. D. P., R. Kusumaardhani, R. A. Rohmawati, P. A. Bhagaskara, I. F. Nafisa, Analysis of waste management in the village of disanah, district of sreseh sampang, madura, JURNAL KESEHATAN LINGKUNGAN 10 (2018) 368. doi:10.20473/jkl.v10i4.2018.368-375.
URL https://e-journal.unair.ac.id/JKL/article/view/6424

[2] K. L. H. dan Kehutanan, Capaian kinerja pengelolaan sampah (2023).

[3] N. Oceanic, A. Administration, What is marine debris? (4 2021).

[4] A. Cherson, The marine debris research, prevention and reduction act: A policy analysis (2005). doi:10.13140/RG.2.2.26619.54562.
URL https://www.researchgate.net/publication/344306086

[5] T. K. N. P. S. Laut, Sampah laut (2023).

[6] T. Hyakudome, Design of autonomous underwater vehicle (2011).
URL www.intechopen.com

[7] Y. C. Chang, S. K. Hsu, C. H. Tsai, Sidescan sonar image processing: Correcting brightness variation and patching gaps, Journal of Marine Science and Technology 18 (2010) 785–789. doi:10.51400/2709-6998.1935.

[8] V. Wiley, T. Lucas, Computer vision and image processing: A paper review, International Journal of Artificial Intelligence Research 2 (2018) 22. doi:10.29099/ijair.v2i1.42.

[9] O. Cosido, A. Iglesias, A. Galvez, R. Catuogno, M. Campi, L. Teran, E. Sainz, Hybridization of convergent photogrammetry, computer vision, and artificial intelligence for digital documentation of cultural heritage-a case study: The magdalena palace, Institute of Electrical and Electronics Engineers Inc., 2014, pp. 369–376. doi:10.1109/CW.2014.58.

[10] Z. Zou, K. Chen, Z. Shi, Y. Guo, J. Ye, Object detection in 20 years: A survey (5 2019), Proceedings of the IEEE 111 (3) (2023) 257–276. doi:10.1109/JPROC.2023.3238524.

[11] L. Du, R. Zhang, X. Wang, Overview of two-stage object detection algorithms, Vol. 1544, Institute of Physics Publishing, 2020. doi:10.1088/1742-6596/1544/1/012033.

[12] M. Carranza-García, J. Torres-Mateo, P. Lara-Benítez, J. García- Gutiérrez,

On the performance of one-stage and two-stage object detectors in autonomous vehicles using camera data, Remote Sensing 13 (2021) 1–23. doi:10.3390/rs13010089.

[13] B. Janakiramaiah, G. Kalyani, A. Karuna, L. V. N. Prasad, M. Krishna, P. V. Potluri, G. K. V. R. Siddhartha, Military object detection in defence using multi-level capsule networks (3 2021).

[14] A. Aralikatti, J. Appalla, S. Kushal, G. S. Naveen, S. Lokesh, B. S. Jayasri, Real-time object detection and face recognition system to assist the visually impaired, Vol. 1706, IOP Publishing Ltd, 2020. doi:10.1088/1742-6596/1706/1/012149.

[15] A. Balasubramaniam, S. Pasricha, Object detection in autonomous vehicles: Status and open challenges (2022).

[16] Y. Hu, G. Liu, Z. Chen, J. Guo, Object detection algorithm for wheeled mobile robot based on an improved yolov4, Applied Sciences 12 (9) (2022). doi:10.3390/app12094769.
URL https://www.mdpi.com/2076-3417/12/9/4769

[17] A. S. Nugraha, Y. Novanto, B. Rahayudi, Supervised virtual-to-real domain adaptation for object detection task using yolo , in: 2024 IEEE Conference on Artificial Intelligence (CAI), IEEE Computer Society, Los Alamitos, CA, USA, 2024, pp. 1359–1364. doi:10.1109/CAI59869.2024.00242
URL https://doi.ieeecomputersociety.org/10.1109/CAI59869.2024.00242

[18] Tibyani, Penerapan region growing pada analisis citra digital untuk pendeteksian sel-sel kanker rahim, Tesis S2 Teknik Elektro (2005).

[19] S. Srivastava, A. V. Divekar, C. Anilkumar, I. Naik, V. Kulkarni, V. Pattabiraman, Comparative analysis of deep learning image detection algorithms, Journal of Big Data 8 (12 2021). doi:10.1186/s40537-021-00434- w.

[20] H. Yang, P. Liu, Y. Hu, J. Fu, , Microsystem Technologies 27 (2021) 1837–1844. doi:10.1007/s00542-019-04694-8.
URL https://link.springer.com/10.1007/s00542-019-04694-8

[21] Y. H. Liao, J. G. Juang, Automatic marine debris inspection, Aerospace 10 (1 2023). doi:10.3390/aerospace10010084.

[22] G. Tata, S.-J. Royer, O. Poirion, J. Lowe, A robotic approach towards quantifying epipelagic bound plastic using deep visual models (5 2021).
URL http://arxiv.org/abs/2105.01882

[23] B. Xue, B. Huang, W. Wei, G. Chen, H. Li, N. Zhao, H. Zhang, An efficient deep-sea debris detection method using deep neural networks, IEEE Journal of Selected Topics in Applied Earth Observations and Remote Sensing 14 (2021) 12348–12360. doi:10.1109/JSTARS.2021.3130238.

[24] A. Algorry, A. Giles Garcia, G. Wolfmann, Real-time object detection and classification of small and similar figures in image processing, 2017, pp. 516–519. doi:10.1109/CSCI.2017.87.

[25] B. Benjdira, T. Khursheed, A. Koubaa, A. Ammar, K. Ouni, Car detection using unmanned aerial vehicles: Comparison between faster r-cnn and yolov3, in: 2019 1st International Conference on Unmanned Vehicle Systems-Oman (UVS), 2019, pp. 1–6. doi:10.1109/UVS.2019.8658300.

[26] M. S. A. B. Rosli, I. S. Isa, M. I. F. Maruzuki, S. N. Sulaiman, I. Ahmad, Underwater animal detection using yolov4, Institute of Electrical and Electronics Engineers Inc., 2021, pp. 158–163. doi:10.1109/ICCSCE52189.2021.9530877.

[27] M. Tian, X. Li, S. Kong, L. Wu, J. Yu, A modified yolov4 detection method for a vision-based underwater garbage cleaning robot, Frontiers of Information Technology & Electronic Engineering 23 (2022) 1217– 1228. doi:10.1631/FITEE.2100473.
URL https://link.springer.com/10.1631/FITEE.2100473

[28] M. Fulton, J. Hong, M. J. Islam, J. Sattar, Robotic detection of marine litter using deep visual detection models, IEEE, 2019, pp. 5752–5758. doi:10.1109/ICRA.2019.8793975.
URL https://ieeexplore.ieee.org/document/8793975/

[29] S. Majchrowska, A. Miko-lajczyk, M. Ferlin, Z. Klawikowska, M. A. Plantykow, A. Kwasigroch, K. Majek, Waste detection in pomerania: non-profit project for detecting waste in environments, Waste Management 138 (2022) 274–284. doi:https://doi.org/10.1016/j.wasman.2021.12.001
URL https://www.sciencedirect.com/science/article/pii/S0956053X21006474

[30] A. S´anchez-Ferrer, J. J. Valero-Mas, A. J. Gallego, J. Calvo- Zaragoza, An experimental study on marine debris location and recognition using object detection, Pattern Recognition Letters (4 2023). doi:10.1016/j.patrec.2022.12.019.

[31] M. Tian, X. Li, S. Kong, L. Wu, J. Yu, Pruning-based yolov4 algorithm for underwater gabage detection, Vol. 2021-July, IEEE Computer Society, 2021, pp. 4008–4013. doi:10.23919/CCC52363.2021.9550592.

[32] N. Tajbakhsh, J. Y. Shin, S. R. Gurudu, R. T. Hurst, C. B. Kendall, M. B. Gotway, J. Liang, Convolutional neural networks for medical image analysis: Full training or fine tuning?, IEEE Transactions on Medical Imaging 35 (2016) 1299–1312. doi:10.1109/TMI.2016.2535302.

[33] T. Diwan, G. Anirudh, J. V. Tembhurne, Object detection using yolo: challenges, architectural successors, datasets and applications, Multimedia Tools and Applications 82 (2023) 9243–9275. doi:10.1007/s11042- 022- 13644-y.

[34] J. Hosang, R. Benenson, B. Schiele, Learning non-maximum suppression , in: 2017 IEEE Conference on Computer Vision and Pattern Recognition (CVPR), IEEE Computer Society, Los Alamitos, CA, USA, 2017, pp. 6469–6477. doi:10.1109/CVPR.2017.685.
URL https://doi.ieeecomputersociety.org/10.1109/CVPR.2017.685

[35] M. Mahasin, I. A. Dewi, Comparison of cspdarknet53, cspresnext-50, and efficientnet-b0 backbones on yolo v4 as object detector, International Journal of Engineering, Science & InformationTechnology 2 (2022) 64–72. doi:10.52088/ijesty.v1i4.291.
URL https://doi.org/10.52088/ijesty.v1i4.291

[36] A. Bochkovskiy, C.-Y. Wang, H.-Y. M. Liao, Yolov4: Optimal speed and accuracy of object detection (4 2020).
URL http://arxiv.org/abs/2004.10934

[37] D. Misra, Mish: A self regularized non-monotonic activation function, in: British Machine Vision Conference, 2020.
URL https://api.semanticscholar.org/CorpusID:221113156

[38] Z. Qiang, W. Yuanyu, Z. Liang, Z. Jin, L. Yu, L. Dandan, Research on real-time reasoning based on jetson tx2 heterogeneous acceleration yolov4, Institute of Electrical and Electronics Engineers Inc., 2021, pp. 455–459. doi:10.1109/ICCCBDA51879.2021.9442515.

[39] S. Liu, L. Qi, H. Qin, J. Shi, J. Jia, Path aggregation network for instance segmentation, 2018, pp. 8759–8768 doi:10.1109/CVPR.2018.00913

[40] J. Redmon, A. Farhadi, Yolov3: An incremental improvement (04 2018). arXiv:1804.02767.

[41] Z. Zheng, P. Wang, W. Liu, J. Li, R. Ye, D. Ren, Distance-iou loss: Faster and better learning for bounding box regression, Vol. 34, 2020, pp. 12993–13000. doi:10.1609/aaai.v34i07.6999.

[42] A. Krizhevsky, I. Sutskever, G. E. Hinton, Imagenet classification with deep convolutional neural networks (2012).
URL http://code.google.com/p/cuda-convnet/

[43] K. Simonyan, A. Zisserman, Very deep convolutional networks for large-scale image recognition (2015). arXiv:1409.1556.

[44] C. Szegedy, W. Liu, Y. Jia, P. Sermanet, S. Reed, D. Anguelov, D. Erhan, V. Vanhoucke, A. Rabinovich, Going deeper with convolutions (2014). arXiv:1409.4842.

[45] O. Ronneberger, P. Fischer, T. Brox, U-net: Convolutional networks for biomedical image segmentation (2015). arXiv:1505.04597.

[46] K. He, G. Gkioxari, P. Doll´ar, R. Girshick, Mask r-cnn, in: 2017 IEEE International Conference on Computer Vision (ICCV), 2017, pp. 2980–2988. doi:10.1109/ICCV.2017.322

[47] M. M. Pasandi, M. Hajabdollahi, N. Karimi, S. Samavi, Modeling of pruning techniques for simplifying deep neural networks, in: 2020 International Conference on Machine Vision and Image Processing (MVIP), 2020, pp. 1–6. doi:10.1109/MVIP49855.2020.9116891.

[48] Z. Liu, J. Li, Z. Shen, G. Huang, S. Yan, C. Zhang, Learning efficient convolutional networks through network slimming (8 2017).
URL http://arxiv.org/abs/1708.06519

[49] K. He, X. Zhang, S. Ren, J. Sun, Deep residual learning for image recognition (2015). arXiv:1512.03385.
URL https://arxiv.org/abs/1512.03385

[50] T. Lin, M. Maire, S. J. Belongie, L. D. Bourdev, R. B. Girshick, J. Hays, P. Perona, D. Ramanan, P. Doll'a r, C. L. Zitnick, Microsoft COCO: common objects in context, CoRR abs/1405.0312 (2014). arXiv:1405.0312.
URL http://arxiv.org/abs/1405.0312